# Deep Structure for end-to-end inverse rendering

Shima Kamyab[1], Ali Ghodsi[2], S. Zohreh Azimifar[1]

[1]School of Electrical and Computer Engineering, Shiraz University, Shiraz, Iran, [2]Department of Statistics and Actuarial Science, University of Waterloo, Waterloo, Canada

*Abstract*— Inverse rendering in a 3D format denoted to recovering the 3D properties of a scene given 2D input image(s) and is typically done using 3D Morphable Model (3DMM) based methods from single view images. These models formulate each face as a weighted combination of some basis vectors extracted from the training data. In this paper a deep framework is proposed in which the coefficients and basis vectors are computed by training an autoencoder network and a Convolutional Neural Network (CNN) simultaneously. The idea is to find a common cause which can be mapped to both the 3D structure and corresponding 2D image using deep networks. The empirical results verify the power of deep framework in finding accurate 3D shapes of human faces from their corresponding 2D images on synthetic datasets of human faces.

*Keywords— Autoencoder, 3D human face Inverse Rendering, Besel face morphable model, Deep 3D human face reconstruction.*

## I. INTRODUCTION

Inverse rendering refers to the set of techniques aimed at retrieving the 3D structure of a particular object class. The goal of inverse rendering is to find the best-suited parameter representation for a 3D object that best describes the structure of a 2D input image [15, 19, 28]. Methods for inverse rendering usually work based on defining a shape space for the desired object class structure and generating new shapes by morphing between existing basis vectors. 3D morphable models (3DMMs) are defined to be such tools for 3D reconstruction. Inverse rendering based on 3DMMs supports obtaining a model for object shape from 2D images and videos, such that their manipulation becomes possible [1 – 8, 27].

In the recent years, advances in 3D sensing techniques such as laser range scanners, have made the analysis of 3D objects more practical. However, usually some visual noise appears in the 3D shapes scanned by these systems. In order to recover the true shape of an object from such noisy data, a shape model can be used that will validate an object's shape by constraining its shape to lie in a defined shape space [1, 2, 9, 10]. Note that we use shape space and shape model in this paper interchangeably

3D shape models are used for many applications such as recognition tasks, virtual avatar control, segmentation of organ shapes in medical images, inverse rendering, etc [2, 11 – 15].

A 3D Shape space is defined based on a number of basis vectors extracted from some existing training data. Principal Component Analyzing (PCA) is the most well-known method for extracting the basis vectors from data that looks for data's maximum variance [15, 20]. A set of coefficients is obtained by projecting each shape onto these bases. The resulting coefficients for each object are called the "shape parameter representation". There should also be some prior knowledge or assumptions, usually formulated as a statistical distribution, describing the data of the desired object class. Using the prior knowledge and the determined basis vectors, new 3D shapes can be generated by weighted combination of the basis vectors. These calculations enables obtaining more accurate representations for 3D shapes in the process of inverse rendering.

Deep learning techniques, originated by the pioneering work of Hinton et. al [29] have recently influenced a large body of intelligent approaches that have demonstrated significant improvements when these methods are applied; it is considered as a powerful approach in many recent vision problems such as inverse rendering systems [7,16,18,21].

A significant volume of research has been devoted to 3D shape retrieval from 2D images in the field of computer vision [2, 3, 6, 13, 15, 18, 19]. Deep learning frameworks have demonstrated superior performance due to their powerful features [7, 16, 17, 18, 19].

The main contribution of the proposed research is to design a learning-based framework for effective and efficient extraction of a 3D human face structure from a single 2D input image. Inspired by existing learning-based systems that use PCA to determine the shape space bases, and to find the best representation of the input image, a 2-part deep network is proposed that extracts the shape space bases in one part and obtains the representation of the input image in the second part. The first part includes an autoencoder [30] that was used for the extraction of 3D basis vectors from training data. Generally, the resulting weights provided by the autoencoder networks have proven to be capable of spanning the space typically spanned by PCA basis vectors [31]. The second part is a deep convolutional neural network (CNN) that is trained to extract the suitable coefficients (representation) for a given 2D input image resulting in its equivalent 3D structure. The 3D structure is computed by a weighted combination of the extracted basis vectors achieved in the first part. These two parts are trained simultaneously using both 2D and 3D training data. Indications from the research validate that the concurrent training strategy helps the autoencoder part to constrain the representation layer to be consistent with 2D training data in the second network. This constraint actually acts as a regularizer for the network to obtain a representation space which causes to find a mapping from both 2D and 3D data to the same representation.

This paper is outlined as follows: Section II presents a literature review of existing methods for 3D human face reconstruction based on deep learning. In Section III, the shape model for human face reconstruction is formulated and discussed. The proposed deep framework is illustrated and analyzed in Section IV. Section V presents the results by comparing the proposed method with some existing, superior reconstruction methods. Section VI provides conclusions and thoughts regarding further developments.

## II. LITERATURE REVIEW ON LEARNING THE SHAPE SPACE FOR INVERSE RENDERING

Statistical 3D shape models for the 3D modeling of human faces was initiated by Blanz and Vetter [20]. This so-called 3D Morphable Model (3DMM) primarily captures the 3D shape, grabs texture information, and then is capable of predicting the 3D face shape and texture from a single 2D input image.

One of the well-known, learning-based methods for inverse rendering of human face was proposed by Aldrian et. al in [15]. This method derives a closed form solution for 3D reconstruction of the human face from a 2D input image using 3DMM by utilizing several assumptions about the image acquisition process and about texture information.

The research field of deep learning has attracted wide attention in recent years due to achievements made in the accuracy and quality of its performance. Accordingly, several research approaches on the 3D structure of various objects' shape have been taken and proposed. In [21] a geometric 3D shape was represented as a probability distribution of binary variables on a 3D voxel grid, using a Convolutional Deep Belief Network (CDBN). This model, called "3D ShapeNets", learns the distribution of complex 3D shapes across different object categories, and arbitrary poses from raw CAD data, and then, discovers hierarchical compositional part representation automatically using a 2.5D depth map as the input.

Xinhan Di and Pengqian Yu In [22] proposed a stacked hierarchical network and an end-to-end training strategy in order to recover 3D structure from a single-view 2D silhouette image. In their research they trained their stacked network for 2 object categories including cars and planes.

In [7] an autoencoder is used for feature learning on 2D images. The input (and output) of the autoencoder is a 2D image obtained by projecting the 3D shape into its corresponding 2D space. Good performance in 3D shape retrieval is reported in this paper shown by aggregating the features learned from 2D images. This proposed method obtained state-of-the-art performance on 3D shape retrieval benchmarks.

In [24, 25] a deep convolutional neural network for 3D human pose and camera parameter estimation from monocular and single 2D images is proposed. Brau and Jiang in [24] designed an additional output layer in the architecture of the network which along with projecting the predicted 3D joint distributions onto 2D distribution, puts some constraints on 3D length of human body parts. Moreover, an independent network is trained in order to learn a 3D pose based on prior distribution. In [25] 3D human pose estimation is proposed using a single raw RGB image that reasons jointly about 2D joint estimation and 3D pose reconstruction to improve these both tasks (i.e. 2D joint estimation and 3D pose reconstruction). This scenario used a multi-stage Convolutional Neural Network (CNN) architecture and the knowledge of plausible 3D landmark locations to refine the search for better 2D locations. The CNN in this work uses probabilistic knowledge of 3D human pose. The reported results show that this method improves state-of-the-art results on the Human 3.6M benchmark in the case of both 2D and 3D errors.

In [26] deep 3D reconstruction techniques are employed for addressing human action recognition. This work considerably improves the state-of-the-art performance and introduces new research directions in the field of 3D action recognition; this is significant following the recent development and widespread use of portable, affordable, high-quality and accurate motion capturing devices (e.g. Microsoft Kinect [35, 36]).

In the study conducted by Richardson et. al. [19], which is the most study similar to the proposed work, a convolutional network is designed for reconstructing a 3D face from a single image in a learning-based approach. The proposed framework contains a few iterations for finding suitable solutions. The training data for this network includes random facial images generated using a 3D face model.

In [32] a deep network with a recurrent component is proposed to obtain novel views from a single input image. This network performs the rotations through a sequence of transformations using the recurrent component.

Wu et al. in [33] used an adversarial structure in their proposed generative deep framework in order to generate 3D object shapes. The adversarial part in their framework consists of a generator for reconstructing the object shapes and a discriminator for controlling the quality of generated shapes by classifying them as real or synthetic. They also used a variational autoencoder in order to extract suitable representations from 2D input images as input to the generator.

In [34] a silhouette based volumetric loss function is proposed for training a convolutional deep network for 3D shape reconstruction from a single 2D input image. The proposed loss function makes it possible to train the network without explicit 3D supervision.

The above mentioned research that proposes deep structures for 3D reconstruction could achieve substantial results. However, there is still potential for improvement in terms of interpretability and quality of the results in the field of deep 3D reconstruction; this is especially true for 3D reconstruction of the human face by reducing artifacts in 3D results and increasing the estimated accuracy. In this paper, a deep method for learning based inverse rendering of the human face is employed that uses synthetic data obtained from Besel Face Model for 3D human face [20] as training data. The output of the network is the weighted combination of some extracted basis vectors which are obtained from the training data. The distribution of the resulting reconstructed faces are controlled and restricted by another deep structure through a parallel training approach in the proposed framework. The proposed framework consists of two parts. One part for extracting basis vectors of the shape space from the training data using an autoencoder. The second part constrains the obtained shape space by the autoencoder to be mapped to one for 2D input images as well. The second part also extracts the suitable shape representation for a 2D input image.

### III. PRELIMINARIES

In order to demonstrate the concepts used in the proposed inverse rendering framework, some basic concepts are described in the following sections, such as shape space modeling and using an autoencoder network to extract their basis vectors.

## A. Defining the shape model

In order to define a shape space, a set of basis vectors and some assumptions about that shape model's properties need to be determined. The basis vectors specify the span, the main structure and the dimensions of the space, and the assumptions constrain the resulting space to introduce feasible shapes for a specific object class. For instance, PCA-based methods include two major modules that respectively determine the direction of largest variations of the data as basis vectors and obtain a diagonal covariance matrix for the projected data on this bases. This approach is actually, in concept, equivalent to estimating a multi-dimensional Gaussian distribution.

Projection of shape vector onto predefined basis functions in the shape space results in a set of coefficients representing the 3D shape. A surface shown by n vertices in $R^3$ is shown by a set of $d$ shape coefficients, forming a vector. A mapping function:

$$F(s): R^d \rightarrow R^{3n} \quad (1)$$

maps the shape from the coefficient (shape) space to the actual 3D space and finding this mapping function is the objective of inverse rendering process. In this paper, we actually assume parametric form for the mapping function by considering a linear shape space and define the generator function F(.) like the one presented in [23] as a weighted combination of the extracted basis functions:

$$F(s) = \overline{F} + \Phi z = \overline{F} + \sum_{i=1}^{d} \Phi_i z_i \quad (2)$$

where $\overline{F}$ is the mean shape obtained from the training data, $\Phi \in R^{3n \times d}$ is a matrix containing basis vectors of the shape space in which $\Phi_i \in R^{3n}$ are the columns and $z \in R^d$ is the vector of shape parameters. The matrix $\Phi$ defines the properties of the shape space and the prior distribution of the training samples.

In the process of defining the parameters of a statistical model corresponding to a human face, if the model consists of a small number of basis vectors, the resulting parameters contain few shape details and the model cannot capture suitable characteristics and variations to represent a promising face. On the other hand, if the number of basis functions is large, the model may overfit the training data. Therefore, the learned space may be biased toward the mean shape. Therefore, the number of dimensions of the resulting shape space should be determined carefully to avoid overfitting or losing information.

The parameters of an existing face shape represent a deformed 2D surface which can be shown by a mesh or point cloud with *n* points. This surface representation can fit to an input data (e.g. 2D image) if there exists a way to measure the quality of fitting like computing the difference between the surface and the input data [2].

By using the shape model described in (2), in order to find the basis vectors of the resulting shape model and parameter representations for a 3D shapes, an autoencoder (AE) network is used in this paper aiming at reconstructing the 3D shape of its input 3D face. In the next section, AE structure and the strategy for the restriction of its latent variables (representations) will be described.

## B. Autoencoder

Traditional autoencoders (AEs) [30] are models (usually multilayer artificial neural networks), designed to output the reconstruction of their input. Specifically, AEs reconstruct input data into hidden representations, and then use these representations to reconstruct outputs that resemble the originals.

An end-to-end AE can be split into two complementary networks: the first part is an encoder, which maps input x to a latent representation or so-called hidden code, say *z*, and the second network is a decoder, that maps the representation to reconstructed input, say *x*. Fig. 1 shows the standard structure of an AE having these to parts.

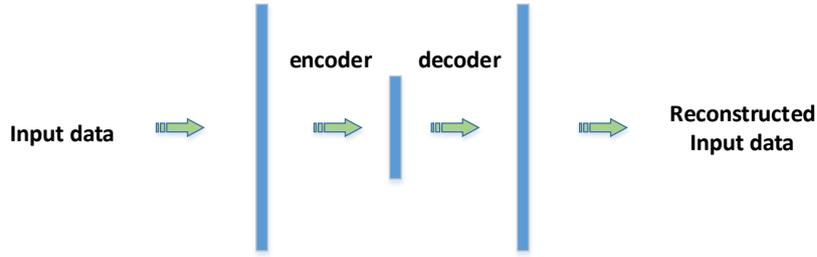

Fig. 1. basic structure of an autoencoder

Using an AE is closely related to performing PCA on the data. Generally, it can be proven that using a standard AE for reconstructing the data actually corresponds to performing PCA [31].

IV. PROPOSED FRAMEWORK

*C. Problem definition*

In order to find suitable parameter values of a statistical shape model for a 2D input image, a cost function should be defined. In order to do this, we first defined a cost function called $J_1$ for finding the suitable parameter values for a 3D shape model in the parameter space as in (3):

$$J_1 = \| x^{(n)} - (\overline{F} + \sum_{i=1}^{d} \Phi_i z_i) \| \quad (3)$$

Where, $\Phi_i$ denotes the $i^{th}$ basis vector of the *d*- dimensional shape space, $\overline{F}$ represents the 3D mean shape as mentioned in previous section and *z* stands for the shape parameter vector. The parenthesis in above formula is actually the reconstruction of $n^{th}$ training input in the shape model and $x^{(n)}$ is the corresponding original 3D training input.

The idea in this paper is to assume a common cause (representation) which results both 3D shape and its corresponding 2D images. We consider this representation as the parameter vector for that shape. Fig. 2 shows this idea as a graphical model.

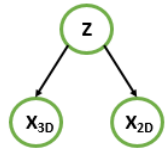

Fig. 2. The common representation which can be mapped to both 3D shape and corresponding 2D input image

In order to apply the above mentioned assumption, if we denote the mappings from 2D and 3D input images as *y*, *z* respectively, we can define another loss function called $J_2$ In order to constrain these two mappings to bring them as close as possible. This loss function can be seen in (4).

$$J_2 = \| z - y \|^2 \quad (4)$$

Where, *z* is the parameter vector described in (3) and y is the mapping of the 2D image to the shape space.

We believe that by minimizing the two cost functions in Equations (3), (4) simultaneously, we can find a suitable parameter representation for a shape consistent with both 2D and 3D data. Therefore, the main minimization problem for finding parameter vector *z* can be defined as (5):

$$z = \arg\min_z(J_1 + J_2) \tag{5}$$

*D. Network architechture*

The general architecture which we used for the proposed network (Fig. 3) consists of two parts which are simultaneously trained and form a main network. In the first part, we used an AE in order to find the shape space for 3D input data. It finds the representations by minimizing the reconstruction error defined in (3). The second part is a convolutional network (CNN) for finding a suitable mapping of the 2D images to the shape space found by the AE in the first part. CNN is a powerful tool to find and extract suitable features and representations from the data. Figure 2 illustrates the main structure of the proposed network during the training phase.

The two parts of the proposed network are related through the loss function $J_1+J_2$ as can be seen in (5). We actually fixed the number of AE's representation neurons to 280 by trial and error considering its effect in achieving promising visual results. The second part which is a CNN, consists of 3 convolutional and 2 feedforward layers. The output of the second network has the same number of neurons as the representation layer of the AE. The representation layer of the AE is related the output of second network through the loss function in (4). These two networks are trained simultaneously. The input to the AE network is actually a vectorized 3D point cloud describing the 3D shape of human face. The input to the second part (CNN), is a tensor describing a single 2D image for each face having some degree of the pose. In the training phase, each 2D input image and its corresponding 3D shape will be feed to each parts in network, simultaneously.

After training the networks using the defined loss function, during the test phase, the second part (CNN) is used alone with the learned basis functions by AE network, to construct 3D face shapes from single 2D images.

*E. Contributions*

The main contributions of the proposed research are as follows:

- The proposed framework provides real-time, fully automated 3D renderings without the need to set landmarks or prior knowledge on the latent variables, as this prior is obtained from the data itself.
- The process in the proposed framework requires only one pass.
- The bases of the defined shape space are learned directly from the 3D data and are consistent with the characteristics of the 2D input images.

## V. EXPERIMENTS

*F. Training data*

In order to provide a suitable collection of 3D scans of human faces for the deep network, the Besel Face Model [20] was used for generating synthetic faces in neutral expressions. About 10000, 3D faces were generated and a $32\times32$ corresponding gray scale 2D input image was captured and rendered for each 3D face as training data to be used in the second network of which 20% were posed up to 90 degrees to right or left. In order to obtain a posed 2D image, the 3D face was rotated and rendered an image from a fixed view. Fig. 4 illustrates some samples in training data with different poses.

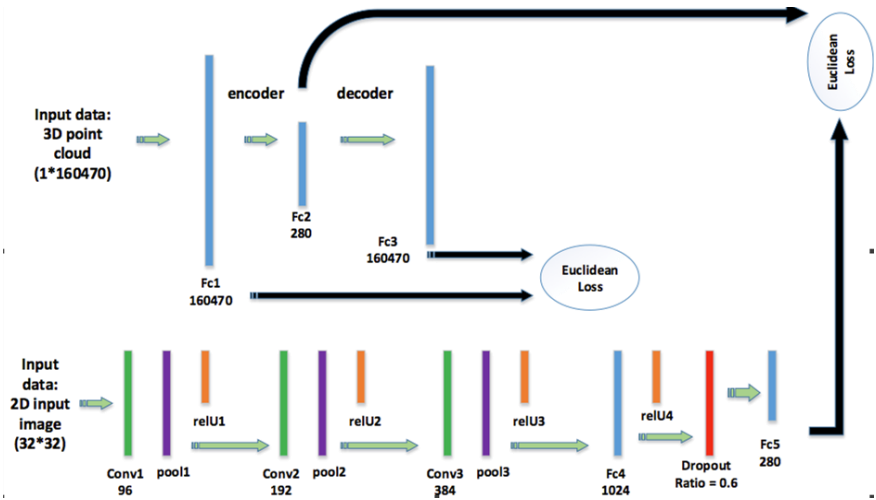

Fig. 3. Proposed network structure

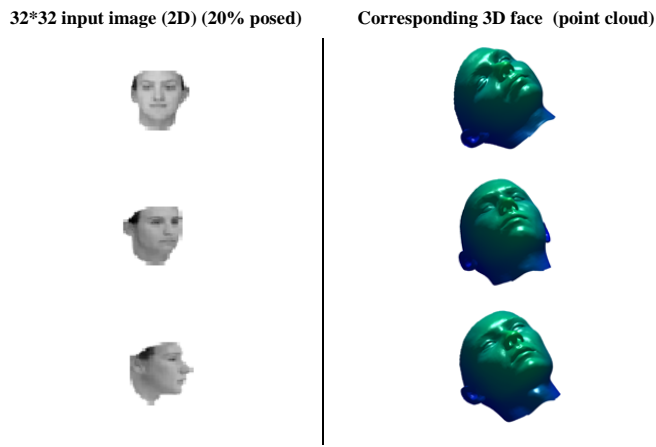

Fig. 4. Sample synthetic training data

Using the Besel Face model, a 3D cloud was obtained with 53490 points (a vector with size equal to 160470 after concatenation of the point coordinates) for each face.

G. *Compared methods*

In this section, the proposed method in this paper is compared to the landmark and PCA-based 3D face inverse rendering method proposed by Aldrian et. al [15], in which, by taking some assumptions about the image formation process and putting some constraints on the results, they proposed a closed form solution for 3D face inverse rendering. They also used the Besel Face Model in their work.

On the other hand, in order to evaluate the effect of using AE in finding the basis vectors for the shape space, we trained our CNN part as a new deep network associated with the BFM basis vectors for 3D human face reconstruction. We reported the compared results using two new measures.

H. *Training setup*

The network was trained on a computer with CPU Corei7, 32 GB Ram, and GTX1080 GPU and Caffe built in Windows 10 was used as the framework for training.

This network was trained for 50,000 iterations with batch size = 60, momentum = 0.9, weight decay = 0.0005, learning rate (fixed police) =10e-06 and Adam solver till average loss reached to 0.7 from 1845.

A dropout layer was used after layer Fc4, with dropout ratio equal to 0.6, in order to regularize the learning. A rectified linear activation function was used after each convolutional layer and feedforward layer, except for the last feedforward layer in the second network.

I. *Visual and Numerical results*

Figure 5 and Table 1 compare the MSE obtained by the proposed network and Aldrian's method visually and numerically. Fig. 5 including visual results, consists of four columns, in which the first column shows the face surface obtained by proposed network for each out of sample face in BFM and the other two columns show the Mean Square Error (MSE) map for compared and proposed method. The MSE map represents the MSE value between each point in ground truth for each face and the result obtained by existing method. Note that in Figure 4, the obtained MSEs are scaled between 0 and 1, in order to ease the comparisons.

Table 1 shows the obtained average MSE for each out of sample face in BFM. In order to compute the MSE in Table 1, and to make the measures comparable with Aldrian's in [15], the network output was scaled so that the new coordinate matched the BFM default and then computed the MSE between the ground truth and predicted face. The Generalized Procrustes analysis (GPA) [37] was also used to align the reconstructed faces to the ground truth face.

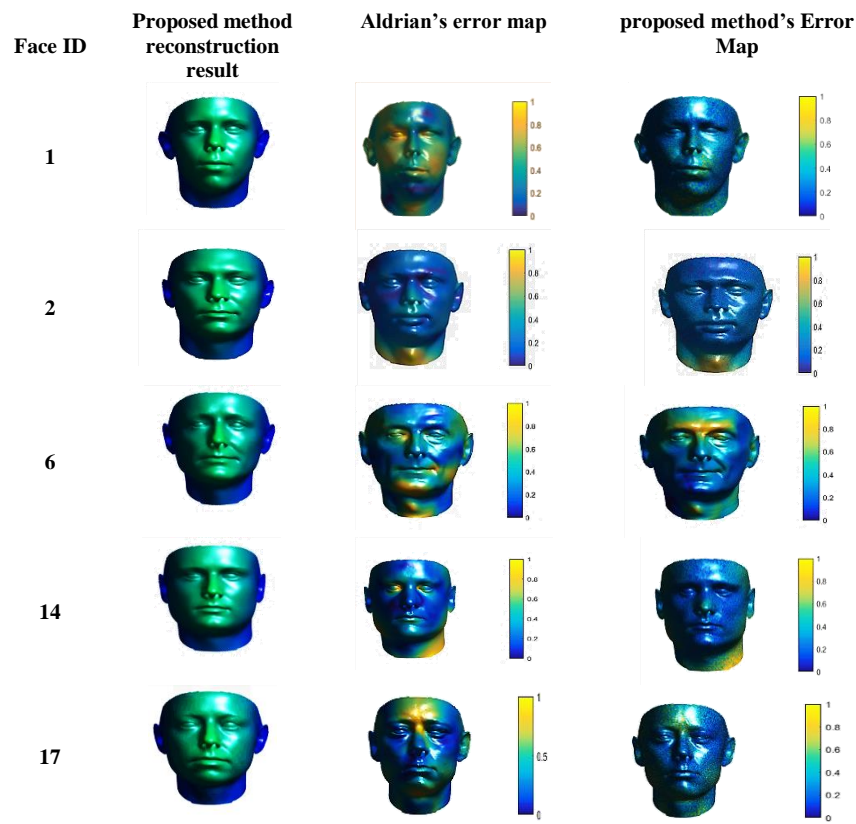

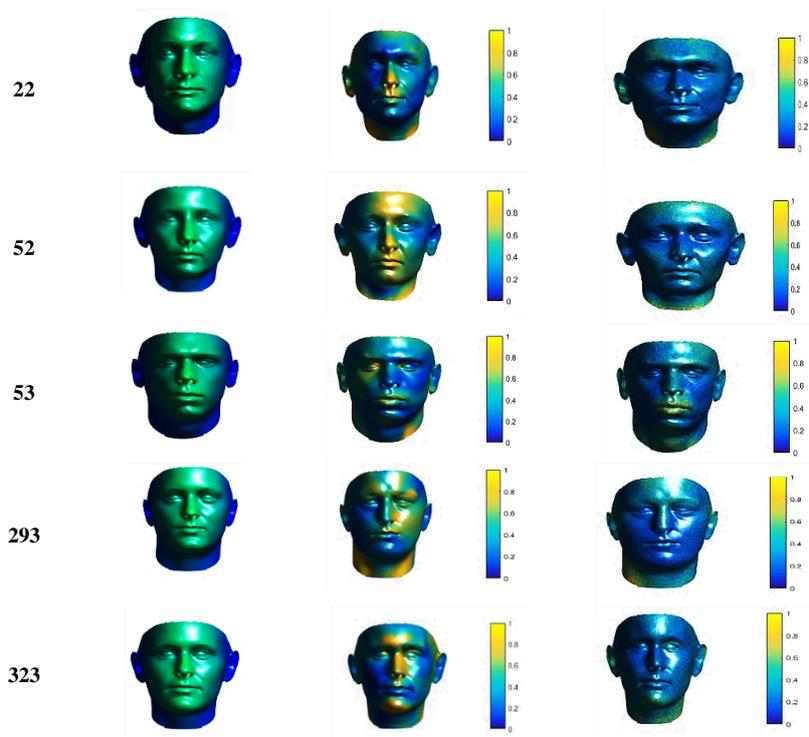

| 22 |
| 52 |
| 53 |
| 293 |
| 323 |

Fig. 5.  MSE map obtained by proposed method and aldrian's method

From Fig. 5 and Table 1 it can be observed that the proposed network, by extracting the basis vectors and coefficients from the data, could achieve better results in most cases.

TABLE I.  MEAN-SQUARED EUCLIDEAN ERROR ($\times 10^{12}$) IN UNITS OF $\mu m^2$

| Face ID | Aldrian's | Proposed network |
|---|---|---|
| 001 | 0.1154 | 0.29852 |
| 002 | 0.2136 | 0.080133 |
| 006 | 1.9752 | 1.4623 |
| 014 | 0.3392 | 0.1667 |
| 017 | 0.2078 | 0.089138 |
| 022 | 1.8462 | 1.3125 |
| 052 | 0.2066 | 0.2726 |
| 053 | 0.1740 | 0.05507 |
| 293 | 0.0779 | 0.089136 |
| 323 | 0.0917 | 0.084962 |
| Mean | 0.5248 | 0.3673 |

The effect of changing the pose in the input image on the output MSE for out of sample faces is illustrated in Figure 6. Aldrian's shape reconstruction method was implemented and its results were compared the results obtained from the proposed method (in a scale like Table 1).

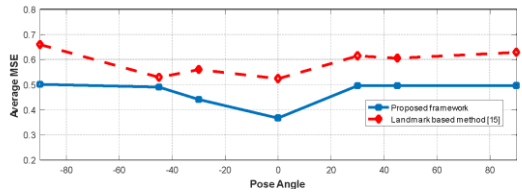

Fig. 6.  Tthe effect of pose angle face in input image on reconstruction error ($\times 10^{12}$ in $\mu m^2$)

*J. Evaluation of extracted basis functions in inverse rendering*

In order to evaluate the power of the extracted basis functions in 3D face reconstruction by the proposed network, its obtained results were compared to a convolutional network trained with the BFM basis vectors for inverse rendering. The compared network structure is actually same as the CNN in the proposed framework and its output CNN is the coefficients by which BFM bases could form a 3D shape.

The setup for training this compared network was the same as the proposed network, but the training data includes the 2D image as the input and the corresponding shape coefficients in Besel Face Model as output for each training data.

We used two measures for comparing the proposed network to new CNN:

(a) Per-vertex normal differences [38] which shows the distribution of differences between the angles of normal vectors of each reconstructed faces with their ground truth for each out of the sample face in BFM. In order to compute the normal vector, called N, for a point in the 3D shape, we considered every 3 adjacent points, called P1, P2, P3, on the face surface and obtained the normal vector of their formed triangle using cross product of their coordinete as in (6).

$$
\begin{aligned}
V &= P2 - P1 \\
W &= P3 - p1 \\
Nx &= (Vy * Wz) - (Vz * Wy) \\
Ny &= (Vz * Wx) - (Vx * Wz) \\
Nz &= (Vx * Wy) - (Vy * Wx)
\end{aligned}
\quad (6)
$$

After obtaining normal vector for each triangle, it will be assigned to each of 3 points and finally the vectors for each point will be outcome of calculated normals for triangles which that point belongs to.

We also computed the difference angle between two normal vectors, called $N_1$ and $N_2$, using cosine formula as in (7).

$$\theta = \arccos\left(\frac{\vec{N}_1 \cdot \vec{N}_2}{\|\vec{N}_1\|\|\vec{N}_2\|}\right) \quad (7)$$

Where, $\theta$ is the angle between two normals and $\|.\|$ denote the magnitude/norm of a vector.

(b) Per-vertex MSE [39] which gives the mean squared error for each vertex of recovered 3D face shape by the networks. In order to compute this measure, the faces were aligned using GPA and then the MSE between the recovered faces and the corresponding ground truth in BFM were computed;

Fig. 7 illustrates the results obtained by the proposed network and the new CNN in terms of two criteria in which the first row includes the comparison using per-vertex differences of normals and the second row shows the results for per-vertex MSE. In this figure, each column corresponds to one of compared methods denoted by the title of that column.

| **Proposed network output** | **CNN trained with BFM** |
|---|---|

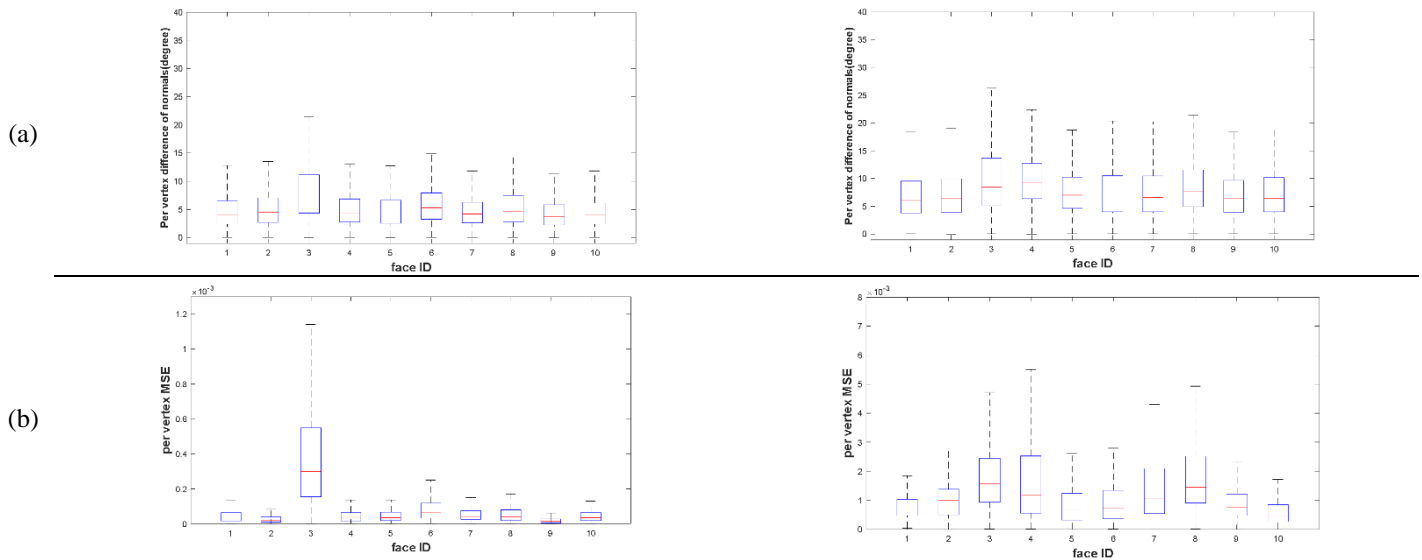

Fig. 7. The result of comparison between proposed network (left column) and the CNN trained and used with BFM (right column): (a) pervertex differences of normals. (b). pervertex MSE between the obtaned face and corresponding ground truth

From Figure 7, it can be observed that the proposed network could achieve better results in terms of computed criteria. We believe these results confirm the effectiveness of extracting the basis vectors from the training data.

## VI. CONCOLUSIONS

In this paper, a deep framework for 3D human face inverse rendering from a single 2D input image is proposed, where the basis vectors for reconstruction of 3D faces are obtained automatically by using an autoencoder, without the need for manual FFP setting on input images. The coefficients for weighted combination of basis vectors are also extracted from the data by designing a CNN in which the 2D input image, is used for training. The two networks are related using a loss function in which the representation layer and the CNN output is constrained to be as close as possible.

The result of comparing the proposed method with the PCA and Landmark based methods shows the effectiveness of proposed network in finding accurate 3D face shapes automatically.

In future works based on the proposed research, we suggest using other types of autoencoder and CNN in order to improve the performance. We also fixed the number of shape space dimensions in our work which can be improved by adaptively tuning in future works.